\documentclass{article}

\PassOptionsToPackage{sort&compress}{natbib}
\usepackage[final]{neurips_2024}
\usepackage{graphicx}
\usepackage{multirow}
\usepackage{amsmath,amsfonts,bm}

\def\eqref#1{equation~\ref{#1}}
\def\1{\bm{1}}

\DeclareMathAlphabet{\mathsfit}{\encodingdefault}{\sfdefault}{m}{sl}
\SetMathAlphabet{\mathsfit}{bold}{\encodingdefault}{\sfdefault}{bx}{n}

\newcommand{\at}[2][]{#1|_{#2}} % Derivative at point

\usepackage[utf8]{inputenc} % allow utf-8 input
\usepackage[T1]{fontenc}    % use 8-bit T1 fonts
\usepackage{amssymb}
\usepackage{amsthm}
\usepackage{amsmath}
\usepackage{bm}
\usepackage{xcolor}
\definecolor{dblue}{HTML}{1D91C0}
\definecolor{dred}{HTML}{D95F0E}
\usepackage[colorlinks=true, linkcolor=dred, citecolor=dblue]{hyperref} 
\usepackage[nameinlink]{cleveref}
\usepackage{url}            % simple URL typesetting
\usepackage{booktabs}       % professional-quality tables
\usepackage{amsfonts}       % blackboard math symbols
\usepackage{nicefrac}       % compact symbols for 1/2, etc.
\usepackage{microtype}      % microtypography
\usepackage{xcolor}         % colors
\usepackage{soul}
\usepackage{comment}
\usepackage{subcaption}
\usepackage{wrapfig}

\usepackage{algorithm}
\usepackage{algpseudocodex}

\creflabelformat{equation}{#2#1#3}

\renewcommand\b[1]{\boldsymbol{#1}}

\newcommand\w{\b{w}}

\newcommand{\data}{\mathcal{D}}
\newcommand{\fim}{\mathcal{I}}

\title{Efficient Model Compression Techniques with FishLeg}
\author{%
  Jamie McGowan$^*$ \\
  MediaTek Research\\
  \And
  Wei Sheng Lai \\
  University College London\\
  \AND
  Weibin Chen \\
  University College London\\
  \And
  Henry Aldridge \\
  University College London\\
  \And
  Jools Clarke \\
  University College London\\
  \And
  Jezabel R Garcia \\
  MediaTek Research\\
  \And
  Rui Xia \\
  University of Cambridge\\
  \And
  Yilei Liang$^\dagger$ \\
  University of Cambridge\\
  \And
  Guillaume Hennequin \\
  MediaTek Research \& University of Cambridge\\
  \And
  Alberto Bernacchia \\
  MediaTek Research\\
}

\begin{document}

\maketitle
\def\thefootnote{$*$}\footnotetext{Correspondence to \texttt{jamie.mcgowan@mtkresearch.com}}
\def\thefootnote{$\dagger$}\footnotetext{Work done while at MediaTek Research.}

\begin{abstract}
In many domains, the most successful AI models tend to be the largest, indeed often too large to be handled by AI players with limited computational resources. To mitigate this, a number of compression methods have been developed, including methods that prune the network down to high sparsity whilst retaining performance. The best-performing pruning techniques are often those that use second-order curvature information (such as an estimate of the Fisher information matrix) to score the importance of each weight and to predict the optimal compensation for weight deletion. However, these methods are difficult to scale to high-dimensional parameter spaces without making heavy approximations. Here, we propose the FishLeg surgeon (FLS), a new second-order pruning method based on the Fisher-Legendre (FishLeg) optimizer. At the heart of FishLeg is a meta-learning approach to amortising the action of the \emph{inverse} FIM, which brings a number of advantages. Firstly, the parameterisation enables the use of flexible tensor factorisation techniques to improve computational and memory efficiency without sacrificing much accuracy, alleviating challenges associated with scalability of most second-order pruning methods. Secondly, directly estimating the inverse FIM leads to less sensitivity to the amplification of stochasticity during inversion, thereby resulting in more precise estimates. Thirdly, our approach also allows for progressive assimilation of the curvature into the parameterization. In the gradual pruning regime, this results in a more efficient estimate refinement as opposed to re-estimation. 
We find that FishLeg achieves higher or comparable performance against two common baselines in the area, most notably in the high sparsity regime when considering a ResNet18 model on CIFAR-10 (84\% accuracy at 95\% sparsity vs 60\% for OBS) and TinyIM (53\% accuracy at 80\% sparsity vs 48\% for OBS).
\end{abstract}

\section{Introduction}
\label{sec:intro}

The current staggering growth of AI models is threatening to sideline small and medium-sized AI contributors with limited access to compute resources, who cannot afford to run the largest models and must therefore compromise on performance.
Large models are also expensive to serve and hard to deploy on low-power devices.
Consequently, there is a growing need for methods that can compress these models down to a fraction of their original size whilst retaining their performance \citep{liu_ten_2023}, or indeed train models from scratch to be sparse~\citep{liu2023seeingbelievingbraininspiredmodular}.

Some of the most promising directions for neural compression revolve around leveraging second order information in order to selectively prune least important parameters, while simultaneously updating those that remain. Several recent studies have shown that second-order parameter importance scores are more accurate than more rudimentary measures derived from weight magnitudes and/or gradients \citep{gale2019state,sanh2020movement}, yielding more effective pruning in convolutional \citep{theis2018faster,singh2020woodfisher} or transformer \citep{kuznedelev2022ovit,kurtic2022optimal} architectures.
Moreover, second-order methods have shown some promise in pruning benchmarks specifically chosen to ``fail current sparse neural networks'' \citep{liu2023sparsity}. However to obtain state-of-the-art performance, compressed models often require a period of retraining after or during the process of model compression which necessitates hand-crafted compression recipes to be designed -- usually switching between compression and training phases~\citep{kuznedelev_ovit_2022}.

Despite the promise of OBS-derived approaches, they are faced with a severe tradeoff between scalability and accuracy that has proven hard to navigate.
Specifically, both the importance scores and the weight updates rely on estimating the action of the inverse Hessian $H^{-1}$ (or, in our case, the inverse Fisher matrix $F^{-1}$) on a high-dimensional parameter space ($\b{v} \mapsto H^{-1} \b{v}$), which inevitably calls for approximations.
Indeed, all recent applications of the OBS framework to pruning have had to make significant simplifications, such as (i) ignoring correlations between most weights or groups of weights \citep{kurtic2022optimal,kuznedelev2022ovit}, even those that belong to the same layer, or (ii) making low-rank approximations to the Hessian \citep{singh2020woodfisher,frantar2021mfac} which are as good as the memory they consume.
Moreover, in the gradual pruning regime where the model changes little from stage to stage, what has been learned about the curvature at the current stage is often unduly discarded in the next one.

In parallel to the advancements of second-order pruning techniques, \emph{optimization} has also been the subject of many improvements that tackle similar computational challenges. In particular, the FishLeg optimizer introduced by \citep{garcia2023fisherlegendre} attacks the scalability-accuracy dilemma by learning to directly amortize $F^{-1} \b{v}$ products in an easy-to-evaluate $Q(\b\lambda) \b{v}$ form.
This is done by minimizing an auxiliary loss $\mathcal{A}(\b\lambda)$ derived from Legendre duality principles, w.r.t.\ a set of auxiliary parameters $\b\lambda$ (details in \Cref{app:fishleg}).
In contrast to low-rank approximations of the Fisher matrix that require hundreds of gradients to be stored, FishLeg allows the progressive distillation of a large number of gradients into the auxiliary parameter set $\b\lambda$. This direct and gradual learning of $F^{-1}$ in $Q(\b\lambda)$ is particularly relevant to the gradual pruning setting, where other methods typically have to recompute $F$ from scratch following pruning, and re-invert it.
By means of low-parameter tensor factorization techniques, the size of $\b\lambda$ can be kept within a small multiple of the size of the model itself, enabling pruning of large models with limited memory.
Whilst such memory efficiency can also be attained through KFAC-based methods~\citep{martens2015optimizing,wang2019eigendamage}, FishLeg's direct estimation of the \emph{inverse} Fisher is less sensitive to gradient noise.
Moreover, the form of KFAC's $F^{-1}$ follows rigidly from approximate mathematical derivations, whereas FishLeg's $Q(\b\lambda)$ can be any user-specified positive-definite quadratic form, yielding greater flexibility and accuracy.
We use this flexibility to develop a novel variation on the well-known Kronecker-factored curvature approximation for dense layers, as well as new approximations for the convolutional layer.

In this work, we introduce the FishLeg Surgeon (FLS) --- a novel pruning algorithm that exploits the inverse curvature estimation machinery of the Fisher-Legendre (FishLeg) optimizer \citep{garcia2023fisherlegendre}. We build on the Optimal Brain Surgeon (OBS; \citealp{lecun1989optimal,hassibi1992second}), a classical approach to pruning that approximates the network's loss function in quadratic form to determine (i) the importance (or saliency) of each weight and (ii) the optimal way of compensating for their deletion.

Our contributions are:
\begin{itemize}
    \item We provide the first example of using a second-order optimizer for unstructured and semi-structured pruning -- allowing for the Fisher matrix to be updated online during pruning/training.
    \item We modify the auxiliary loss $\mathcal{A}(\b\lambda)$ to facilitate assessment of its convergence and to promote learning of the full $F^{-1}$ as required for pruning (as opposed to learning the action of $F^{-1}$ on the subspace of momentary noisy gradients, as relevant to the optimization setting of \citealp{garcia2023fisherlegendre}).
    \item We propose a new preconditioner for this (often ill-conditioned) auxiliary loss, and show analytically that it accelerates convergence asymptotically.
    \item We propose a new initialization scheme for $Q(\b\lambda)$ that leads to better estimation of $F^{-1}$ especially when it is ill-conditioned.
    \item We evaluate our proposed method on CIFAR-10 \citealp{krizhevsky2009learning} and TinyIM \citealp{le2015tiny} datasets with a ResNet18 model in the unstructured and N:M semi-structured pruning regimes and compare these against readily available baselines.
\end{itemize}

\section{Neural Compression with FishLeg}
\label{sec:theory}

\begin{algorithm}[h]
    \caption{FishLeg Surgeon (FLS)}\label{alg:gradualprune}
    \begin{algorithmic}[1]
        \State \textbf{Goal}: gradual pruning to $f_{\text{end}}$\% sparsity.
        \State Choose hyperparameters: damping factor $\gamma$, learning rate $\eta$, Adam parameters, sparsity schedule $\{ f_t \}$.
        \State Pretrain $Q(\b\lambda_0)$. \Comment{starts with a good estimate of the inverse FIM}
        \State $t \leftarrow 0$, $\b w_0 \leftarrow \b w^*$
        \While{not finished}
            \LComment{Pruning step}
            \State Select and prune the $(f_{t+1}-f_t)\%$ least important weights using $\b w_{t+1} = \b w_{t}^{2} / \text{diag}(F_{\gamma}^{-1}(\b w_t))$ and the latest approximation $F^{-1}_{\gamma}(\b w_t) \approx Q(\b\lambda_t)$, where $\b w_t$ is the masked parameters from the previous sparsity level.
        \For{$s=1:S$} 
            \State $\mathcal{L}, \b g\leftarrow$ value and gradient of loss evaluated at the masked $\tilde{\b w}_{s}$ on a data minibatch
            \LComment{Fine-tuning}
            \State $\tilde{\b w}_{s+1} \leftarrow \tilde{\b w}_{s} - \eta[Q(\tilde{\b\lambda}_s)\b g]$ \Comment{masked update that preserves current sparsity}
            \LComment{Update the inverse FIM approximation, taking into account the new parameters}
            \State Perform one step of auxiliary loss minimization, yielding a new $\tilde{\b\lambda}_{s+1}$.
        \EndFor
        \LComment{resume pruning with the fine-tuned parameters and updated inverse FIM estimate}
        \State $\b{w}_{t+1} \leftarrow \tilde{\b{w}}_S, \b{\lambda}_{t+1} \leftarrow \tilde{\b{\lambda}}_S, t \leftarrow t+1$
        
        \EndWhile
    \end{algorithmic}
\end{algorithm}

Under the standard assumption that the gradient at the current point $\mathbf{w}$ is negligible for a pretrained model, the OBS formulas for the optimal weight to be pruned $w_p$ and the corresponding update $\delta_p$ can be derived by writing the locally quadratic problem under the constraint that element $p$ of $\delta_p$ is equal to $-w_p$, which means that $w_p$ is zero after applying the update to $\mathbf{w}$. This problem has the following Lagrangian:
\begin{equation}
L(\delta_p, \lambda) = \delta_p^\top \mathbf{H} \delta_p + \lambda (\mathbf{e}_p^\top \delta_p - (-w_p)), \quad (6)
\end{equation}
where $\mathbf{H}$ denotes the Hessian at $\mathbf{w}$ and $\mathbf{e}_p$ is the $p$-th canonical basis vector. 

The inverse FIM used to score and update the weights is typically recomputed at regular intervals in current second order pruning methods. The reason for this being that OBS derived methods are based around inverting an estimation of the empirical Fisher. However during pruning, as parameters are removed and others updated, the inverse FIM estimation becomes increasingly inaccurate, necessitating a complete re-estimation of the empirical Fisher before its explicit inversion.\footnote{Indeed, much of the advancements of OBS methods revolve around efficient inversion techniques.}

Here, we reason that FishLeg's parametric estimation of the inverse FIM, $Q(\b\lambda)$, can be actively updated in a rolling fashion between consecutive pruning steps by simply performing a certain number of auxiliary loss minimization steps.
Crucially, by amortizing the re-computation of the inverse FIM in this way, we can afford to update our model directly (for which we use the FishLeg optimizer, also based on the running estimate $Q(\b\lambda)$), as outlined in \Cref{alg:gradualprune}.
Hence, unlike previous approaches to gradual second-order pruning, we need not re-estimate and re-invert the Fisher matrix from scratch after each pruning step -- we simply refine our current estimate.

\subsection{Memory efficient parameterization of the inverse Fisher approximation}
\label{subsec:layers}

For scalability, we approximate $F^{-1}$ in block-diagonal form, with each layer contributing one block.
Note that these blocks are orders of magnitude larger than the ones used in previous second-order approaches that implemented direct inversion (e.g.\ \citealp{kurtic2022optimal} used blocks of size 50).

Our choice of structure for $Q(\b\lambda)$ is slightly more constrained by our pruning objective than it is for the FishLeg optimizer: we require efficient evaluation of not only $Q\b{v}$ products (which essentially sketch the curvature in the direction of steepest descent), but also $\text{diag}(Q)$ (required in \Cref{eq:obs_score}).
For dense layers with $n_\text{i}$ inputs and $n_\text{o}$ outputs, we parameterize the corresponding inverse Fisher block as
\begin{equation}
   \label{eq:block_linear}
   Q(\b\lambda) \triangleq D (LL^\top \otimes RR^\top) D
\end{equation}
where $L \in \mathbb{R}^{n_\text{o} \times n_\text{o}}$ and $R \in \mathbb{R}^{n_\text{i} \times n_\text{i}}$ are two parameter matrices, $D$ is a diagonal matrix with $(n_\text{i}+1) n_\text{o}$ parameters, and $\otimes$ denotes the Kronecker product. 
This construction is such that, for $V \in \mathbb{R}^{n_\text{o} \times n_\text{i}}$, 
\begin{equation}
Q(\b\lambda) \text{vec}(V) = D \odot \text{vec}(LL^\top (V \odot \bar{D}) RR^\top)
\end{equation}
with the (unusual) convention that $\text{vec}(\cdot)$ vectorizes row-wise (corresponding to a no-copy reshape in numerical code), and $\odot$ denotes elementwise (Hadamard) product.  
Here, $\bar{D} \in \mathbb{R}^{n_\text{o} \times (n_\text{i}+1)}$ is the un-vectorized version of the diagonal of $D$.
Similarly, 
\begin{equation}
\text{diag}(Q) = \text{diag}(D)^2 \odot (\text{diag}(LL^\top) \otimes \text{diag}(RR^\top))
\end{equation}
can be evaluated efficiently, with $\text{diag}(LL^\top) = (L \odot L) (1, \ldots, 1)^\top$.
Note that the inclusion of $D$ makes it more expressive than the standard KFAC approximation which is limited to the Kronecker product. 
For completeness in \Cref{app:ablation}, we compare the above parameterisation with a pure diagonal parameterisation and also a more restrictive block diagonal structure similar to other second-order pruning methods (i.e. oBERT \& M-FAC).

For convolutional layers, we follow a similar tensor factorization strategy.
Filter parameters are tensors of dimensions $n_\text{o} (\text{output channels}) \times n_\text{i} (\text{input channels}) \times K (\text{kernel size})$.
Whilst we could parameterize the inverse Fisher block as a 3-way Kronecker product, \citet{grosse2016kronecker}'s KFAC derivation for convolutional layers suggests combining together the input and kernel-size dimensions.
We therefore use the same structure as in \Cref{eq:block_linear}, but with $R$ of size $n_\text{i} K$ and $D$ of size $n_\text{o} n_\text{i} K$.

\subsection{Initialization of $Q$}
\label{subsec:qinit}

Our experiments with FishLeg have revealed that the minimization of the auxiliary loss is very sensitive to initialization -- to the point that getting it wrong can yield useless estimates of $F_\gamma^{-1}$.
In the context of neural network optimization, \citet{garcia2023fisherlegendre} advocated an identity initialization $Q_0 = \alpha I$.
To choose the value of $\alpha$, they observed that this identity initialization implied that the FishLeg update $\w_{t+1} \leftarrow \w_t - \eta Q(\b\lambda) \nabla_{\w}\mathcal{L}$  would initially correspond to SGD.
Thus, given a learning rate $\eta_\text{SGD}$ known to work well for SGD, they set $\alpha \triangleq \eta_\text{SGD} / \eta$.
However, in the context of pruning this rationale no longer applies; we therefore revisited the choice of $\alpha$.
\begin{figure}[t]
    \centering
    \includegraphics[width=0.9\textwidth]{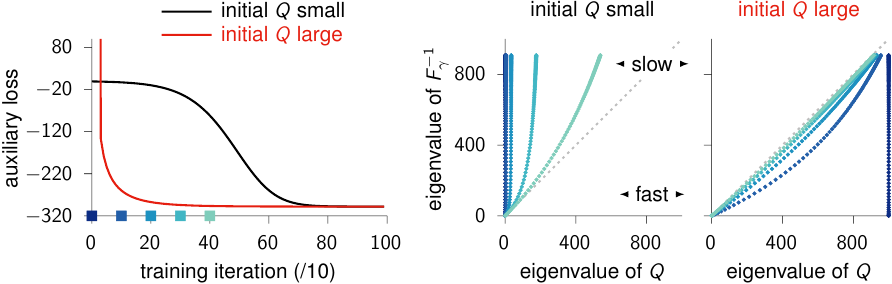}
    \caption{\textbf{The initialization of $Q(\b\lambda)$ matters much.}
    In this toy experiment, the true Fisher matrix ($n=100$) was chosen so that its $i^\text{th}$ eigenvalue is $\xi_i \triangleq 1/i^2$, and the damping parameter $\gamma$ was set to $10^{-3}$. Thus, the eigenvalues of $F_\gamma^{-1}$ lie roughly in the $[1 - 1000]$ range.
    The auxiliary loss $\mathcal{A}(Q) = \frac12 \text{Tr}(QFQ) - \text{Tr}(Q)$ (left) was minimized by gradient descent w.r.t.\ the Cholesky factor of $Q(\b\lambda)$, initialized such that $Q(\b\lambda)=I$ (black) or $Q(\b\lambda)=\gamma^{-1} I = 1000 \times I$ (red).
    The learning rate was optimized separately for each case.
    This simulation shows that it is clearly better to initialize $Q$ to be large rather than small.
    Indeed, a simple derivation shows that  each eigenvalue $\beta_i$ of $Q$ approaches its target $1/(\xi_i + \gamma)$ at a speed proportional to $(\xi_i + \gamma)$ (\Cref{eq:aux_dynamics}).
    In other words, the eigenvalues of $Q$ that must end up large are also those that evolve the slowest.
    It, therefore makes sense to initialize them to be large so they have less to travel; the eigenvalues that must end up small will become small rapidly anyway.
    The right panels illustrate this behaviour by plotting the eigenvalues of $Q$ against their respective targets, at regular intervals during optimization (color-coded), for both initialization schemes.
    The auxiliary loss is minimized when $\beta_i = 1/(\xi_i+\gamma)$, i.e.\ when the dots lie along the identity line (dashed grey).
    }
    \label{fig:aux_illustration}
\end{figure}

We found that good pruning results could only be obtained for sufficiently large $\alpha$.
To understand this, we studied the idealized dynamics of auxiliary loss gradient descent (\Cref{fig:aux_illustration}; see also \Cref{app:precond}).
Let $F = U \Xi U^\top$ be the eigendecomposition of the Fisher matrix, with $\Xi = \text{diag}(\xi_1, \ldots, \xi_n)$.
Assuming $\b{u} \sim \mathcal{N}(0, I_n)$, the auxiliary loss (\Cref{eq:aux}) reduces to $A(\b\lambda) = \frac12 \text{Tr}(Q F_\gamma Q) - \text{Tr}(Q))$.
Expressing $Q$ in the eigenbasis of $F$ as $Q = U \beta U^\top$, the gradient flow for this deterministic loss function takes the form $\dot{\beta} = - (\Xi + \gamma I) \beta + I$ with $\beta(0) = \alpha I$.
It is then easy to see that $\beta$ will remain diagonal throughout, and that the $i^\text{th}$ eigenvalue of $Q$ has the following dynamics:
\begin{equation}
   \label{eq:aux_dynamics}
   \underbrace{(\xi_i + \gamma)^{-1}}_{\text{time constant}} \frac{d\beta_i}{dt} = - \beta_i + \underbrace{(\xi_i + \gamma)^{-1}}_{\text{optimal steady state}}
   \quad \text{with} \quad
   \beta_i(0) = \alpha.
\end{equation}
Thus, the eigenvalues of $Q$ -- all initially equal to $\alpha$ -- converge at very different speeds depending on their optimal steady states: eigenvalues that must reach large (resp.\ small) values evolve slowly (resp.\ fast).
We therefore conclude that a good initialization is to set $\alpha$ to be as large as the largest eigenvalues of $F_\gamma^{-1}$, namely $(\min\{\xi_i\} + \gamma)^{-1} \approx \gamma^{-1}$.
This way, the eigenvalues of $Q$ that would normally slowly evolve towards $\gamma^{-1}$ are positioned there from the outset, and the eigenvalues that are set to decrease do so rapidly.
\Cref{fig:aux_illustration} illustrates this behaviour and shows that large initialization of $Q$ (with $\alpha \approx \gamma^{-1}$) results in faster minimization of the auxiliary loss. 

\subsection{Preconditioning of the auxiliary loss}
\label{subsec:precond}

Learning the full $F^{-1}$ is a hard problem when $F$ is ill-conditioned, as the auxiliary loss inherits this ill-conditioning.
Our theoretical analysis of this problem (\Cref{app:precond}) has led to the discovery of a good preconditioner which only costs a single additional $Q(\b\lambda)v$ product per iteration. This preconditioner greatly accelerates asymptotic convergence of the auxiliary loss (\Cref{fig:simple}A), leading to better estimates of the inverse FIM.

\section{Empirical Investigations}
\label{sec:experiments}

In order to validate our approach for pruning and training in the same breath using a second-order optimizer such as FishLeg, we provide initial studies with a ResNet18 model with a single-layer classification layer on CIFAR-10 and TinyIM. We consider two pruning approaches that are common in the field: \textit{unstructured} and \textit{semi-structured} pruning. Across all experiments, we consider the same dense models trained with Adam to 83.4\% test accuracy on CIFAR-10 and 55.4\% test accuracy on TinyIM. Results are reported as mean over 3 random seeds. All experiments were run on a single NVIDIA GeForce RTX 2080 GPU with 8GB of VRAM.

\begin{figure}[t]
    \centering
    \includegraphics[width=0.49\textwidth]{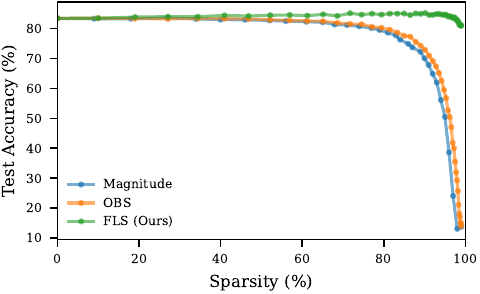}
    \includegraphics[width=0.49\textwidth]{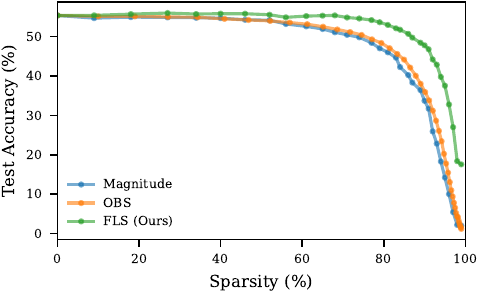}
    \caption{Test accuracy as a function of model sparsity for ResNet18 on CIFAR-10 (left) and TinyIM (right). Different pruning frameworks are used, which are magnitude pruning (blue), OBS (orange) and FLS (green).}
    \label{fig:unstr_comp}
\end{figure}

In addition to the results presented in this section, in \Cref{app:ablation} we provide extensive ablation studies for the methods introduced in \Cref{sec:theory} -- including a direct comparison between the block diagonal approximation used in OBS methods across multiple block sizes.

\subsection{Unstructured Pruning}\label{subsec:unstr}

Unstructured pruning consists of choosing parameters across the entire network to prune and achieve some non-uniformly sparse model which maintains performance on a task. In these experiments, the non-zero parameters were fine-tuned for 1 epoch (with SGDm for magnitude and OBS) after each pruning step following a exponential sparsity schedule. \Cref{fig:unstr_comp} displays a marked improvement over magnitude and OBS pruning methods across both CIFAR-10 and TinyIM experiments -- with only a reduction of $\sim -2.5\%$ from dense performance at 99\% sparsity on CIFAR-10. In addition to this, FLS benefits from being able to continually adapt its second-order information between pruning steps. By amortizing the overhead in recomputing the empirical FIM we afford efficient second-order updates during the finetuning phases -- which, combined with some potential effect from Occam's razor~\citep{blumer1987occam}, explains the slight increase in performance across the experiment.

\subsection{Semi-Structured Pruning}\label{subsec:semi}

\begin{wrapfigure}[11]{R}{0.5\textwidth}
    \centering
    \begin{tabular}{l l l}
        \hline
        \multirow{ 2}{*}{Method} & \multicolumn{2}{c}{Test Accuracy (\%)} \\
         &  CIFAR-10 & TinyIM \\
        \hline
        Magnitude & 80.28 & 47.53 \\
        OBS & 81.50 & \textbf{52.04} \\
        FLS (Ours) & \textbf{84.03} & 51.09 \\
        \hline
    \end{tabular}
    \caption{2:4 semi-structured pruning performance of ResNet18 model finetuned on CIFAR-10 and TinyIM data.}
\end{wrapfigure}

Semi-structured (N:M) pruning involves pruning N parameters in each block of M parameters. This repeated pattern can be readily applied to Sparse Tensor Cores which perform calculations on a compressed version of a sparse matrix. Here we consider the setting of 2:4 sparsity which results in a 50\% sparse network. Similar to the unstructured setting, all models were pruned with an exponential schedule up to a 2:4 pattern and retrained for 1 epoch after each pruning step. Our method achieves greater or similar performance when compared to the baselines, and even improves upon the dense model performance in the case of CIFAR-10.

\section{Discussion, limitations and future work}
\label{sec:discussion}

We have introduced a new perspective on second-order pruning that blurs the lines between second-order optimization and compression. We have identified challenges with the naive approach of using an ``optimization" sketch of the FIM for compression and addressed these with expanded parameterisations, which we have justified theoretically and thorough ablation studies. In addition, we have provided empirical evidence which demonstrates our method on ResNet18 with CIFAR-10 and TinyIM.

Despite this, pruning with FishLeg has several limitations. One of the key assumptions in our approach is that the inverse Fisher $F_\gamma^{-1}(\w^\star)$ can be well approximated by a specific form of positive definite matrix $Q(\b\lambda)$; however, the structure chosen for $Q$ is largely dictated by scalability requirements, and may not be appropriate under certain conditions. 
We have proposed memory-efficient factorizations of $Q$ which we have found effective for dense and convolutional layers, and we leave the development of other types of neural network layers to future research. Indeed, it remains to be seen whether these results scale to larger models, other compression techniques (i.e., fully structured, quantisation, a mixed setting etc.) and to a wider variety of layers.
We hope that further research in these areas will likely extend and refine the capabilities of the proposed method.

\clearpage
\bibliographystyle{apalike}
\bibliography{neurips_2024}
\clearpage
\appendix

\section{Fisher-Legendre (FishLeg) Optimizer}
\label{app:fishleg}
FishLeg \citep{garcia2023fisherlegendre} is a scalable second-order optimizer that approximates the natural gradient $F^{-1} \nabla_{\w} \ell(\w, \data)$ based on the following insights.
Let $\w$ be a fixed set of model parameters.
Consider the regularized cross entropy between $p(\data|\w)$ and $p(\data|\w+\b\delta)$,
\begin{equation}
  \label{eq:kdef0}
  \mathcal{H}_\gamma(\b\delta) =
  \mathbb{E}_{\data \sim p(\data|\w)} \ell(\w+\b\delta,\data) + \frac{\gamma}2  \| \b\delta \|^2,
\end{equation}
where $\gamma>0$ is a small damping parameter.
The Legendre-Fenchel conjugate of $\mathcal{H}_\gamma(\b\delta)$ is defined as
\begin{equation}
  \label{eq:legendre}
  \mathcal{H}^\star_\gamma(\b{u}) \triangleq
  \underset{\b\delta}{\text{min}} \
  \mathcal{H}_\gamma(\b\delta)
  - \b{u}^\top \b\delta
  \quad
  \text{with minimizer denoted by }
  \tilde{\b\delta}_\gamma(\b{u}).
\end{equation}
\citeauthor{garcia2023fisherlegendre} were able to prove that, if the negative log-likelihood $\ell(\b{w},\data) = -\log p(\data|\b{w})$ is twice differentiable, then the inverse damped Fisher information matrix exists and is equal to
\begin{align}
  \label{eq:invfish}
  F_\gamma^{-1} \triangleq [F + \gamma I]^{-1} = \nabla_{\b{u}}\tilde{\b\delta}_\gamma(\b{0}).
\end{align}
FishLeg meta-learns a parametric approximation $\overline{\b\delta}(\b{u}, \b\lambda)$ of $\tilde{\b\delta}_\gamma(\b{u})$, by minimizing the auxiliary loss $\mathcal{A}(\b\lambda, \b{u}) \triangleq \mathcal{H}_\gamma(\overline{\b\delta}(\b{u}, \b\lambda))
  - \b{u}^\top \overline{\b\delta}(\b{u}, \b\lambda)$ w.r.t.\ meta-parameters $\b\lambda$, as prescribed by \Cref{eq:legendre}.
Importantly, \Cref{eq:invfish} shows that one only needs to learn the \emph{local} behaviour of the vector field $\tilde{\b\delta}_\gamma(\b{u})$ around small $\b{u}$; thus, \citeauthor{garcia2023fisherlegendre} directly parameterized its (symmetric, positive definite) Jacobian $Q(\b\lambda) $ at $\b{u}=\b{0}$, corresponding to the choice $\overline{\b\delta}(\b{u}, \b\lambda) \triangleq Q(\b\lambda) \b{u}$.
Furthermore, considering the limit of small $\b{u}$ and averaging over a relevant distribution (more on this below and in \Cref{app:aux_dev}), the auxiliary loss becomes
\begin{equation}
  \label{eq:aux}
  \mathcal{A}(\b\lambda) \triangleq
  \mathbb{E}_{\b{u}} \left\{ \frac1{\| \b{u} \|^2} \left[
    \frac12 \b{u}^\top Q(\b\lambda) F_\gamma Q(\b\lambda) \b{u}
    - \b{u}^\top Q(\b\lambda) \b{u} \right] \right\}
\end{equation}
which can be estimated and differentiated efficiently in a number of ways (details in \Cref{sec:theory}).

Practical note: as $Q(\b\lambda)$ converges towards $F_\gamma^{-1}$, the auxiliary loss as defined by \Cref{eq:aux} converges towards $\left\langle - \b{u}^\top F_\gamma^{-1} \b{u}/ \| \b{u} \|^2 \right\rangle$, which is problem-dependent; this makes it hard to assess the quality of our inverse Fisher estimation.
We therefore assess convergence by computing a slightly modified auxiliary loss where we drop the $\frac12$ factor; this should converge to zero. 

Taking the gradient of \Cref{eq:aux} w.r.t.\ $\b\lambda$ makes is clear that $Q(\b\lambda)$ will learn to approximate the action of $F_\gamma^{-1}$ on the subspace spanned by the $\b{u}$'s.
Given their application to natural gradient optimization, \citeauthor{garcia2023fisherlegendre} took those $\b{u}$'s to be stochastic gradients of the model's primary loss function.
For our pruning purposes, however, \Cref{eq:obs} suggests that we must accurately estimate the action of $F$ on the entire parameter space; we will therefore work with a more isotropic distribution of $\b{u}$ (\Cref{sec:theory}).

Directly estimating the inverse Fisher matrix, and doing so in this way, brings a number of advantages.
First, the FishLeg approach is flexible: one can specify any form of $Q(\b\lambda)$, and in particular combine structured approximations obtained through mathematical derivations (as in e.g.\ KFAC; \citealp{martens2015optimizing,grosse2016kronecker,george2018fast}) with a variety of parametric adjustments for greater expressiveness.
We give examples of such choices in \Cref{subsec:layers}.
Second, the FishLeg approach is less biased than KFAC and related methods.
These methods start by assuming that $F$ has a certain structure (e.g.\ block diagonal), obtain a good approximation of $F$ conforming to this structure, and then invert it.
One expects both systematic errors as well as stochasticity in the estimate of $F$ to propagate to $F^{-1}$.
In contrast, FishLeg `fits' a parametric approximation to $F^{-1}$ directly, conveniently avoiding inversion.
Relatedly, a key property of \Cref{eq:aux} is that it is \emph{not} biased by stochasticity in the estimate of $F_\gamma$ (\Cref{app:estimation}; \Cref{fig:estimation}) -- unlike other seemingly sensible auxiliary loss functions such as $\mathbb{E}_{\b{u}} \| Q(\b\lambda) \hat{F}_\gamma \b{u} - \b{u} \|^2$ or $\mathbb{E}_{\b{u}} \| \hat{F}_\gamma Q(\b\lambda) \b{u} - \b{u} \|^2$ whose quadratic terms in $\hat{F}_\gamma$ do survive averaging.

\section{Auxiliary loss derivation}
\label{app:aux_dev}

Starting from the auxiliary loss definition given in \Cref{eq:aux} and in Equation 15 of \cite{garcia2023fisherlegendre}, we can expand the first term with a Taylor Expansion as:
\begin{equation}
    \mathcal{H}_\gamma(\overline{\b\delta}(\b{u}, \b\lambda)) = \mathcal{H}_\gamma(\b{0}) + \nabla_{\boldsymbol{\bar{\delta}}}\mathcal{H}_\gamma(\boldsymbol{\theta},\boldsymbol{\overline{\delta}})\at{\boldsymbol{\bar{\delta}} = \bf 0}\boldsymbol{\overline{\delta}} + \frac{1}{2}\boldsymbol{\overline{\delta}}^\top\nabla^2_{\boldsymbol{\bar{\delta}}}\mathcal{H}_\gamma(\boldsymbol{\theta},\boldsymbol{\overline{\delta}})\at{\boldsymbol{\bar{\delta}} = \bf 0}\boldsymbol{\overline{\delta}}.
\end{equation}

As stated in Appendix A.2 of \cite{garcia2023fisherlegendre}, each term in this Taylor expansion can be expressed as:
\begin{eqnarray}
&&\nabla_{\boldsymbol{\delta}}\mathcal{H}_\gamma(\boldsymbol{\theta},\boldsymbol{\delta})\at{\boldsymbol{\delta} = \bf 0} =
    \mathbb{E}_{\data \sim p(\data|\boldsymbol{\theta})} \nabla_{\boldsymbol{\theta}}\ell(\boldsymbol{\theta},\data) + {\bf 0} ={\bf 0}\\
&&\nabla^2_{\boldsymbol{\delta}}\mathcal{H}_\gamma(\boldsymbol{\theta},\boldsymbol{\delta})\at{\boldsymbol{\delta} = \bf 0}=\mathbb{E}_{\data \sim p(\data|\boldsymbol{\theta})} \nabla^2_{\boldsymbol{\theta}}\ell(\boldsymbol{\theta},\data) + \gamma I =\fim(\boldsymbol{\theta}) + \gamma I = F_{\gamma}.
\end{eqnarray}
where the $0^\text{th}$ order term follows from the fact that we define the minimum at $\boldsymbol{\delta} = \bf 0$, the $1^\text{st}$ order term is zero since we are at a minimum and the $2^\text{nd}$ order term characterizes the Fisher information matrix.

Using the above definitions, one can arrive at,
\begin{equation}
  \mathcal{A}(\b\lambda, \b{u}) =
    \frac12 \overline{\b\delta}(\b{u}, \b\lambda)^\top F_\gamma \overline{\b\delta}(\b{u}, \b\lambda) -
    \b{u}^\top \overline{\b\delta}(\b{u}, \b\lambda) 
\end{equation}
where the second term in \Cref{eq:aux} is unchanged.

\section{Analysis of FishLeg's auxiliary loss \& preconditioning}
\label{app:precond}

In this section, we analyze the minimization dynamics of a generalized version of FishLeg's auxiliary loss:
\begin{equation}
   \mathcal{A}(Q) = \langle \frac12 \b{u}^\top Q^\top P F_\gamma Q \b{u} - \b{u}^\top Q^\top P \b{u} \rangle_{\b{u} \sim \mathcal{N}(0,I)}   
\end{equation}
where $F$ is the model's (damped) Fisher information matrix,  $P$ is a symmetric positive definite matrix, and $Q$ is our approximation of $F_\gamma^{-1}$.
For simplicity, we will assume that the parameterization of $Q$ is non-limiting, i.e.\ we will consider the minimization of $\mathcal{A}$ directly as a function of $Q$.

This loss can be evaluated analytically:
\begin{align}
   \mathcal{A}(Q) &= 
   \left\langle
   \text{Tr}\left(
    \frac12 Q^\top P F_\gamma Q \b{u}\b{u}^\top 
    - Q^\top P \b{u} \b{u}^\top
   \right)
   \right\rangle_{\b{u} \sim \mathcal{N}(0,I)} \\
   &=
   \text{Tr}\left[
    \left(\frac12 Q^\top P F_\gamma Q 
    - Q^\top P 
    \right)
    \left\langle
   \b{u} \b{u}^\top 
   \right\rangle_{\b{u} \sim \mathcal{N}(0,I)} 
   \right] \\
   &=
   \text{Tr}\left( \frac12 Q^\top  P F Q
   - Q^\top P \right)
\end{align}
The optimal $Q^\star$ must satisfy 
\begin{equation}
   0 = \left.\frac{\partial \mathcal{A}}{\partial Q}\right|_{Q = Q^\star}
   = P (F Q^\star - I)
\end{equation}
Therefore, if $P$ and $F_\gamma$ are both invertible, then $Q^\star = F_\gamma^{-1}$ as desired.
To understand how quickly $Q$ will converge to this solution, it is useful to analyze the gradient flow
\begin{equation}
    \frac{dQ}{dt} = - P(F_\gamma Q(t) - I)
\end{equation}
with initial condition $Q(0)= \alpha I$.
Let  $F = U \Lambda U^\top$ be the eigendecomposition of the Fisher matrix, with $\Lambda = \text{diagm}(\lambda_1, \ldots, \lambda_n)$ and $U^\top U = U U^\top = I$.
We will assume that $P$ has the same eigenvectors as $F$, i.e. $P = U \text{diagm}(p_1, \ldots, p_n) U^\top$. 
Rewriting the above gradient flow in the eigenbasis of $F$, we obtain
\begin{align}
   \frac{d}{dt}(U^\top Q(t) U) &=
    - U^\top P (F_\gamma Q - I) U \\
    &= -U^\top U \text{diagm}(p_1, \ldots, p_n) U^\top
    (U (\Lambda + \gamma I) U^\top Q - I) U \\
    &= - \text{diagm}(p_1, \ldots, p_n)
    ((\Lambda + \gamma I) U^\top Q U - I)  \label{eq:q_flow}
\end{align}
We see that if $U^\top Q U$ is diagonal at time $t$, it will remain diagonal.
Given that $U^\top Q U = U^\top (\alpha I) U = \alpha I$ is diagonal, we conclude that at any time $t$,   $U^\top Q(t) U = \text{diagm}(\beta_1(t), \ldots, \beta_n(t))$.
Thus, \Cref{eq:q_flow} boils down to a set of $n$ decoupled, scalar flows,
\begin{equation}
    \frac{d\beta_i}{dt} = -p_i \left[(\lambda_i + \gamma) \beta_i - 1\right]
    \quad \text{with } \beta_i(0) = \alpha
\end{equation}
These equations are more easily interpreted when rewritten as
\begin{equation}
   \frac{\beta_i^\star}{p_i} \frac{d\beta_i}{dt}
   = -\beta_i + \beta_i^\star
\end{equation}
where $\beta_i^\star = (\lambda_i + \gamma)^{-1}$ is the corresponding eigenvalue of the solution $Q^\star$ (the ``target eigenvalues'').
The solution to these dynamics is
\begin{equation}
    \beta_i(t) = \beta_i^\star
     + (\alpha - \beta_i^\star) \exp\left(
     \frac{-t}{\tau_i}
     \right)
     \quad \text{with }
     \tau_i \triangleq \frac{\beta_i^\star}{p_i}.
     \label{eq:q_flow_1D}
\end{equation}
For $P = I$, i.e. $p_i = 1$, we recover the result of the main text (c.f.\ \Cref{fig:aux_illustration}):
$\beta_i$ converges exponentially to its target $\beta_i^\star$, but on a timescale $\tau_i$ proportional to $\beta_i^\star$ itself.
This is a problem when $F_\gamma$ is poorly conditioned, such that there is a broad range of $\beta_i^\star$: in this case, some $\beta_i$'s will converge rapidly, and some others will converge very slowly.

\Cref{eq:q_flow_1D} suggests a solution based on a judicious choice of the preconditioner $P$.
If somehow we could precondition the loss with $P = F_\gamma^{-1}$, then $p_i = \beta_i^\star$ and therefore $\tau_i = 1$ for all $i$ -- this case we have rapid uniform convergence of the inverse Fisher in all directions.  
While we do not know $F_\gamma^{-1}$ (indeed this is what we are trying to learn \ldots), we do know that $Q(t)$ is supposed to converge (albeit slowly) towards $F_\gamma^{-1}$.
Thus, we propose a simple time-dependent preconditioner $P(t) = Q(t)$.
Empirically, we do find that this choice leads to better asymptotic convergence of the auxiliary loss, as illustrated in \Cref{fig:simple}A.
Note that this only costs a single additional $Q \b{v}$ product in every iteration.

\section{FishLeg inverse curvature estimation: flexible and accurate}
\label{app:estimation}

\begin{figure}[t]
    \centering
    \includegraphics[width=\textwidth]{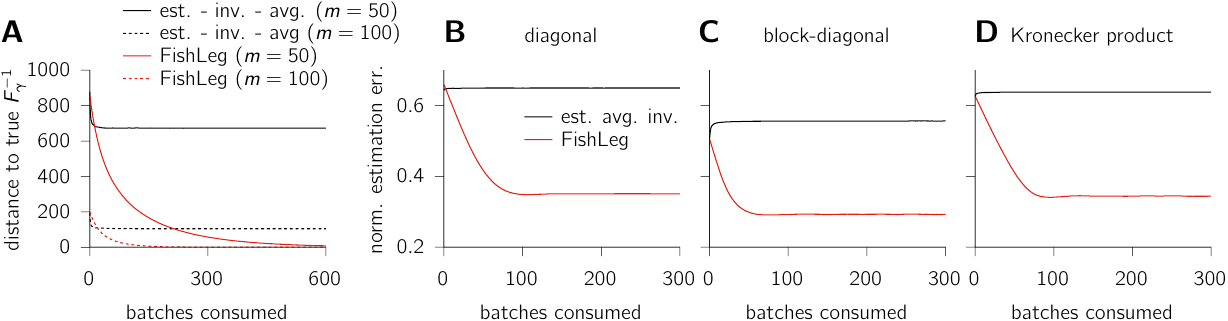}
    \caption{%
    \textbf{Assessing FishLeg's inverse curvature estimation in a controlled setting}. 
    In this figure, the true Fisher matrix $F \in \mathbb{R}^{100\times 100}$ is constructed to have a random orthonormal eigenbasis and eigenvalues $\lambda_i \propto e^{-i/30}$. All results are averaged over 20 independent realizations of the corresponding experiment with different random seeds.
    \textbf{(A)}: standard affine-invariant Riemannian distance between FishLeg's $Q$ and $F_\gamma^{-1}$ ($\gamma=0.01$), as a function of the number of data mini-batches of size $m$ consumed so far.
    Each Adam step of auxiliary loss minimization consumes one minibatch.
    In this case, we use a full parameterization $Q = LL^\top$ that contains the solution $F^{-1}_\gamma$; in that case, FishLeg's inverse curvature estimation is consistent and the error goes to zero. 
    As a baseline, we show the behaviour of a simple but biased estimator that estimates $F_\gamma$ on each new minibatch, inverts that noisy estimate, and averages the result over minibatches; inverting noisy estimates yields a bias that persists asymptotically.
    \textbf{(B-D)}: In these panels, the inverse Fisher is estimated in structured form (B: diagonal; C: block-diagonal, 5 blocks; D: Kronecker product, $(5\times 5) \otimes (20 \times 20)$.
    This is done either by FishLeg assuming a correspondingly structured form for $Q$ (red), or by (i) approximating $F_\gamma$ in structured form for each minibatch (for the Kronecker approximation, we use a permuted SVD to find the nearest Kronecker product in the least-squares sense; \cite{van1993approximation}), (ii) averaging over minibatches (for the Kronecker approximation the two factors are averaged separately, as in KFAC), and (iii) inverting the result (black; note that in this case, the inverse inherits the structure).
    We report the squared error between $Q\b{u}$ and $F_\gamma^{-1} \b{u}$, averaged over $\b{u} \sim \mathcal{N}(0, \Sigma_u)$, and normalized by the average norm of $F_\gamma^{-1} \b{u}$. Here, to reflect the need of accurately estimating the action of $F^{-1}_\gamma$ on the least salient parameter dimensions, we have chosen $\Sigma_u = F^{-1}$.
    \label{fig:estimation}
    }
\end{figure}

In this section, we report on a series of simple experiments that show that FishLeg's inverse curvature estimation is typically more accurate and flexible than more conventional approaches.

First, \Cref{fig:estimation}A shows that -- when the parameterization of FishLeg's $Q$ is sufficiently expressive to include $F^{-1}_\gamma$, $Q$ converges to $F_\gamma^{-1}$ as desired, despite only having access to stochastic estimates of $F$. 
This is because, using standard unbiased estimates of the Fisher matrix (or, practically, Fisher-vector products) on mini-batches in \Cref{eq:aux}, FishLeg's auxiliary loss and its gradient are also unbiased.
With sufficiently small learning rate, we therefore expect $Q$ to converge to the inverse damped Fisher solution.
In contrast, a more naive scheme that computes an average of inverses of noisy Fisher estimates (`est. -- inv. -- avg.' in \Cref{fig:estimation}A) yields a bias that persists asymptotically.

Second, when $F_\gamma^{-1}$ lies outside the domain of the structured approximation (e.g.\ when it is not exactly a single Kronecker product, or a block-diagonal matrix), there is an advantage to directly approximating $F_\gamma^{-1}$ in the desired structured form $Q$ (FishLeg's strategy), rather than approximating $F$ in such a form and then inverting the result.
For one, \citep{garcia2023fisherlegendre} had already argued that the former is more flexible than the latter, because one can use structured forms that need not be easily inverted (indeed FishLeg does not invert anything).
Here, we show that even when the structured form is easily inverted, FishLeg still has a marked advantage (\Cref{fig:estimation}B-D). In particular, the auxiliary loss allows the specification of a distribution of vectors $\b{u}$ (specifically, their covariance) to promote learning the action of $F^{-1}_\gamma$ on select directions in parameter space.
This is not possible in a more conventional approach whereby the Fisher matrix is first approximated in structured form, then averaged, and finally inverted.

\section{Second-order Pruning: OBS-based methods}
\label{app:sec_prune}
Most second-order pruning methods are based on the Optimal Brain Surgeon (OBS; \citealp{hassibi1992second}).
OBS begins with a quadratic approximation of the loss function around the pre-trained parameter set $\w^\star$, typically assumed to be a minimum of the loss,
\begin{equation}
  \delta\mathcal{L}(\delta \w)
  \quad \triangleq \quad
  \mathcal{L}(\w^\star + \delta\w) - \mathcal{L}(\w^\star)
  \quad
  \approx
  \quad
  \frac12 \delta\w^\top H(\w^\star) \delta\w,
\end{equation}
where $H(\w^\star)$ is the Hessian of the loss at $\w^\star$.
Here, we will approximate the Hessian by the Fisher $F(\w^\star)$; most other works use the empirical Fisher matrix instead.
This quadratic approximation leads to an analytical solution to the problem of optimally compensating for the deletion of a given weight $w_i$:
\begin{equation}
  \label{eq:obs}
  \delta \w^\star = - \frac{w^\star_i}{[F^{-1}(\w^\star)]_{ii}}
  F^{-1}(\w^\star) \b{e}_i
\end{equation}
where $\b{e}_i$ is the $i^\text{th}$ canonical basis vector~\citep{hassibi1992second}.
The corresponding (minimal) increase in loss resulting from the deletion of weight $w_i$ is taken as its importance score:
\begin{equation}
  \label{eq:obs_score}
  \rho_i =
  \frac{w_i^2}{2 [F^{-1}(\w^\star)]_{ii}}.
\end{equation}
These equations have also been extended 
to handle the semi-structured pruning setting whereby small blocks of weights are treated as single units \citep{kurtic2022optimal}.

Existing second-order pruning methods mostly differ in the way they estimate
$F^{-1} \b{v}$ products to compute \Cref{eq:obs,eq:obs_score}.
All scalable methods make a block-diagonal approximation for $F$.
WoodFisher \citep{singh2020woodfisher} and oBERT \citep{kurtic2022optimal} partition the parameter space into small blocks assumed to be independent, and use the Woodbury identity to recursively update an estimate of the inverse empirical Fisher $\hat{F}^{-1}_{\mathcal{B}}$ for each block $\mathcal{B}$.
These approaches have substantial memory requirements ($\mathcal{O}(|\mathcal{B}| n)$, where $|\mathcal{B}|$ is the block size and $n$ is the total number of parameters in the model).
M-FAC \citep{frantar2021mfac} modifies this recursion to operate directly on $\hat{F}^{-1}_\mathcal{B} \b{v}$ products, in a way that obviates the need for storing $\hat{F}^{-1}_\mathcal{B}$ (some parts of the computation can be cached and reused for any $\b{v}$).
This is typically much slower but requires less memory.
In our work, FLS too approximates $F^{-1}$ in block-diagonal form, but with much larger blocks corresponding to entire layers, and with blocks structured to guarantee computational and memory efficiency.

\section{Ablation Experiments}
\label{app:ablation}

\begin{figure}[!h]
    \centering
    \includegraphics[width=\textwidth]{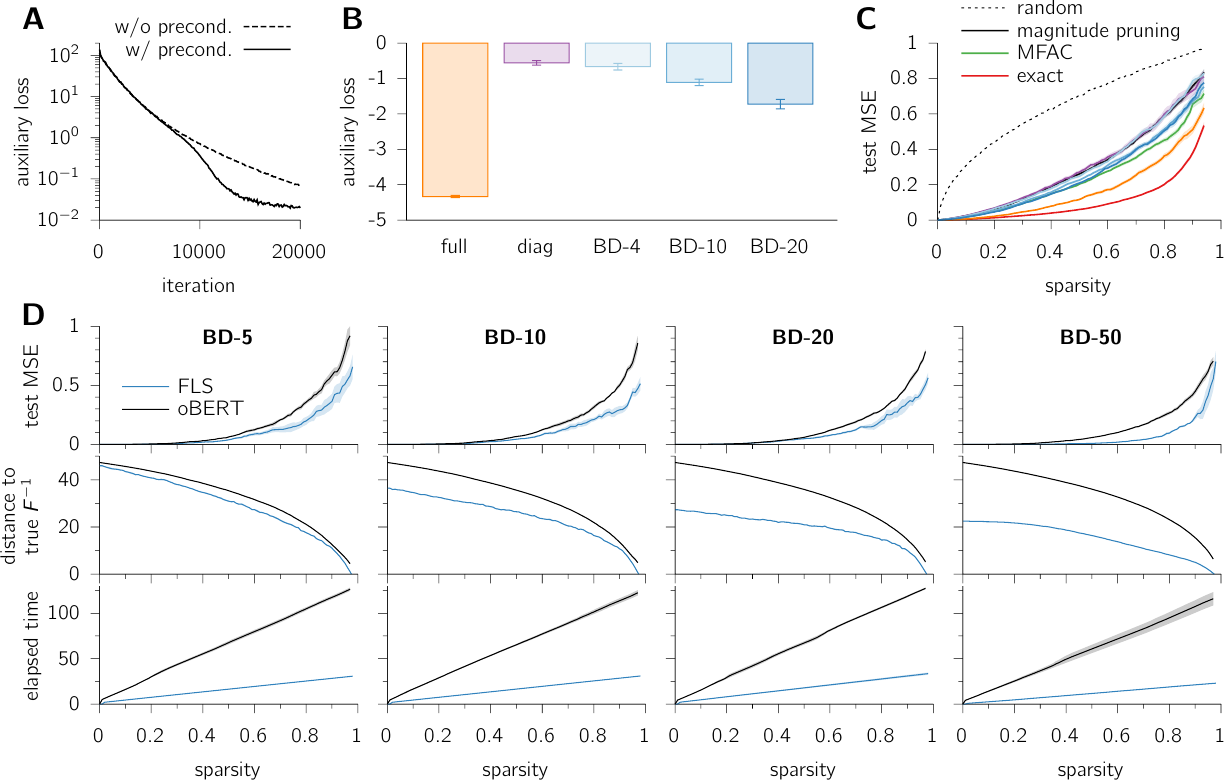}
    \caption{%
    \textbf{Ablation experiments on synthetic data} in a toy setup to show: (A) the utility of preconditioning the auxiliary loss, (B) the predicted quality of the approximated Fisher in different scenario's, (C) the one-shot pruning performance of various Fisher approximations (including other baselines) and (D) the effect of implementing a block diagonal FishLeg approximation and it's comparison to oBERT (an OBS-derived approach) at various block sizes.
    \label{fig:simple}
    }
\end{figure}
For the experiments discussed in this section, a simple linear layer with $n$ inputs and a single output is used to perform controlled ablations and compare various approximations of the inverse Fisher and their impact on one-shot pruning. In \Cref{fig:simple}A-C we choose $n=100$ and in \Cref{fig:simple}D we set $n=500$. The layer weights are drawn from $\mathcal{N}(0, 1/n)$, and inputs are drawn from $\mathcal{N}(0, \Sigma_x)$, where $\Sigma_x$ is a random covariance matrix with eigenvalues $\{ \lambda_i \propto e^{-i/10} \}$. Results are reported as mean $\pm$ s.e.m. over random seeds. Across all experiments, a batch size of $100$ is chosen along with a damping parameter $\gamma = 0.01$. Note that in this toy example, the Fisher matrix is $F = \Sigma_x$, and does not depend on the weights.
Figure~\ref{fig:simple}A shows the effect of preconditioning the FishLeg auxiliary loss using the momentary approximation $Q(\b{\lambda})$ of the inverse Fisher matrix. We observe that this preconditioning does indeed lead to faster asymptotic convergence. This is shown here for the `full' approximation $Q = LL^\top$, which -- in this case -- is as expressive as the Kronecker parameterization of dense layers we have used in the experiments from the main text. 

Figure~\ref{fig:simple}B displays the quality of approximation of the inverse damped Fisher matrix, as measured by FishLeg's auxiliary loss after convergence\footnote{Where the Adam learning rate is separately tuned for each approximation.}, for various parameterizations of $Q(\b\lambda)$. We compare the `full' parameterization $Q = LL^\top$ (orange), a positive diagonal parameterization (purple), and a set of positive-definite block-diagonal approximations with various block sizes (blues). These results show very clearly that a full approximation can achieve a much lower auxiliary loss when compared to less powerful approximations in this case.

Following from this, \Cref{fig:simple}C is reporting the one-shot pruning performance (test MSE) for the various FishLeg parameterizations shown in \Cref{fig:simple}B, as well as for magnitude pruning (black), MFAC ($m=10$; green) and `exact FLS' with $F = \Sigma_x$ appropriately masked and inverted before each pruning step (red). One can observe that the full approximation achieves a far closer performance to the `exact' result across all other baselines in this study. Note that in this case, the `exact FLS' characterises the limit of performance for second-order pruning methods. In this setting, we therefore find a strong correlation between the quality of the iFIM approximation (as measured by \cite{garcia2023fisherlegendre}'s auxiliary loss after convergence) and one-shot pruning performance (comparing \Cref{fig:simple}B and \Cref{fig:simple}C). In particular, block-diagonal approximations (as used by OBS/oBERT) perform worse than the Kronecker-factored approximation (in this case also exact) and, indeed not much better than magnitude pruning or a simple diagonal approximation of the iFIM. Likewise, FLS with a Kronecker-factored Q performs better than MFAC (with rank parameter $m$ generously set to 10, i.e. 10\% of the parameter count, which would normally be intractable memory-wise).

Finally, \Cref{fig:simple}D provides a comparison between FLS with block-diagonal parameterization and oBERT for various block sizes (5, 10, 20, 50). In particular, this ablation study shows benefits of directly estimating the inverse FIM than estimating the FIM and inverting it. oBERT utilizies the WSM formula for effective estimation without explicit inversion, resulting in iterative update of the inverse of moving average for the empricial Fisher matrix. In the top panels, we present one-shot pruning performance (test MSE) as a function of sparsity for the two methods. In the middle panels, the standard affine-invariant Riemannian distance between the masked approximate block-diagonal inverse and the true masked Fisher inverse are shown, for each method. In the bottom panels, the wall-clock time as a function of sparsity is shown. For these experiments, oBERT uses $512$ gradients at each pruning step, whereas FLS performs $20$ steps of auxiliary loss minimization between pruning updates. These results show a systematic improvement in the inverse FIM estimates when using FLS, which implies that directly approximating the inverse Fisher in block-diagonal form (FLS) is better than approximating the Fisher in block-diagonal form before inverting each block (oBERT).

\section{Additional Experimental Details}
\label{app:hyperparameters}
Across all experiments we used a batch size of 128 and additionally applied standard flipping and cropping augmentations. 
\Cref{tab:final-hyperpatameters} show the hyperparameter values used for each of the experimental setups, for the FishLeg optimizer. For other methods, we used an implementation of SGDm (with learning rate set at $1e-3$ and a momentum value of $0.9$ and all others set at the PyTorch SGDm default) which preserved the sparsity map. More details of the experiments can be found in \Cref{sec:experiments}.

\begin{table}[ht]
\centering
\begin{tabular}{ccc}
\hline
Hyperparameter & CIFAR-10 & TinyIM \\
\hline 
Batch Size & 128 & 2048 \\
% \hline
$\eta$ & $10^{-3}$ & $10^{-2}$ \\
% \hline
$\alpha$ & $10^{-5}$ & $10^{-5}$ \\
% \hline
$\eta_\text{aux}$ & $10^{-5}$ & $10^{-6}$ \\
% \hline
$\beta$ & 0.9 & 0.9 \\
% \hline
Damping $\gamma$ & $10^{-3}$ & 1.0 \\
% \hline
Scale & 10 & 1 \\
% \hline
Warmup & 100 & 100 \\
\hline
\end{tabular}
\caption{\label{tab:final-hyperpatameters}
Optimal hyperparameter values for FishLeg, identified as the result of a grid search. These hyperparameters were chosen to minimise the training loss across pruning. Any parameters not shown are left as default values in the FishLeg optimizer library.}
\end{table}

\end{document}